\documentclass[letterpaper, 10 pt, conference]{ieeeconf}  

\IEEEoverridecommandlockouts                              

\usepackage{cite}
\usepackage{amsmath}
\usepackage{multirow}
\usepackage{graphicx}
\usepackage{epstopdf}
\usepackage{epsfig}
\usepackage{bm}
\usepackage{pifont}
\usepackage{verbatim}
\usepackage{bigstrut,multirow,rotating}
\usepackage{booktabs}
\usepackage{multicol}
\usepackage{color}
\usepackage{mathrsfs}
\usepackage{textcomp}
\usepackage[]{algorithm2e}
\usepackage{mathrsfs,soul}

\makeatletter

\newcommand{\Rmnum}[1]{\expandafter\@slowromancap\romannumeral #1@}
\makeatother

\IEEEpeerreviewmaketitle
\nocite{*}

\title{\LARGE \bf
Grasping Detection Network with Uncertainty Estimation for Confidence-Driven Semi-Supervised Domain Adaptation}

\author{Haiyue~Zhu$^{1}$, Yiting~Li$^{2}$, Fengjun Bai$^{3}$, Wenjie~Chen$^{4}$, \\ Xiaocong~Li$^{1}$, Jun~Ma$^{2}$, Chek~Sing~Teo$^{1}$, Pey Yuen Tao$^{1}$, and Wei~Lin$^{1}$
\thanks{$^{1}$H.~Zhu, X.~Li, C.~S.~Teo, P.~Y.~Tao, and W.~Lin are with the Mechatronics Group, Singapore Institute of Manufacturing Technology (SIMTech), Agency for Science, Technology and Research, Singapore 638075 (e-mail: \{zhu\_haiyue, li\_xiaocong, csteo, pytao, wlin\}@simtech.a-star.edu.sg).}
\thanks{$^{2}$Y.~Li and J.~Ma are with the Department of Electrical and Computer Engineering, National University of Singapore, Singapore 117583 (e-mail: yiting\_li@u.nus.edu, elemj@nus.edu.sg).}
\thanks{$^{3}$F.~Bai is with the Advanced Robotics Application Group, Advanced Remanufacturing and Technology Centre (ARTC), Agency for Science, Technology and Research, Singapore 637143 (e-mail: bai\_fengjun@artc.a-star.edu.sg).}
\thanks{$^{4}$W.~Chen is with the School of Electrical Engineering and Automation, Anhui University, Hefei, China 230601 (e-mail: wjiechen@yahoo.com.sg).}
}

\begin{document}

\maketitle
\thispagestyle{empty}
\pagestyle{empty}

\begin{abstract}
Data-efficient domain adaptation with only a few labelled data is desired for many robotic applications, e.g., in grasping detection, the inference skill learned from a grasping dataset is not universal enough to directly apply on various other daily/industrial applications. This paper presents an approach enabling the easy domain adaptation through a novel grasping detection network with confidence-driven semi-supervised learning, where these two components deeply interact with each other. The proposed grasping detection network specially provides a prediction uncertainty estimation mechanism by leveraging on Feature Pyramid Network (FPN), and the mean-teacher semi-supervised learning utilizes such uncertainty information to emphasizing the consistency loss only for those unlabelled data with high confidence, which we referred it as the confidence-driven mean teacher. This approach largely prevents the student model to learn the incorrect/harmful information from the consistency loss, which speeds up the learning progress and improves the model accuracy. Our results show that the proposed network can achieve high success rate on the Cornell grasping dataset, and for domain adaptation with very limited data, the confidence-driven mean teacher outperforms the original mean teacher and direct training by more than 10\% in evaluation loss especially for avoiding the overfitting and model diverging.

\end{abstract}

\section{Introduction}

Grasping detection is a fundamental problem as most of the sequential robotic manipulation is heavily relying on the successful grasping, e.g. sorting, assembly, etc. The goal for the grasping detection is to find a proper grasp configuration (location and pose), so that the object can be firmly grasped by the gripper. Traditionally, the grasping detection is commonly achieved by human-designed features or 2D/3D model evaluation~\cite{SAHBANI2012326,Bohg2014}, which calculates the optimal grasping configuration for some certain quality metrics. However, such methods are often tedious and not applicable to novel objects, especially in unstructured environments.

With the recent advances in machine learning, the grasping detection is more and more fulfilled by deep learning~\cite{Chu2018,Umar_ijcai2018} to directly learn the grasping candidates from the trials or human reasoning. The strong generalization capability of deep learning enables such detection networks to tackle well with the novel unseen objects. However, the deep learning for robotic grasping detection is a bit different from the standard computer vision problems as it is quite domain-related and may not be very general. The inference skill acquired from some public datasets may not be directly applicable to other real applications due to the difference in grasping background, sensor, view angle, etc. The traditional solution is to prepare a new domain dataset with full annotations and then apply transfer learning for a direct domain adaptation. However, data preparation and labelling might be quite tedious, especially for the application that may change frequently. As a result, a simple and efficient domain adaptation approach with less labelled data is desired for the robotic grasping application to fill the last-mile gap between artificial intelligence and real-world applications.

\begin{figure}[tb!]
\centering{\includegraphics[width=3.0in]{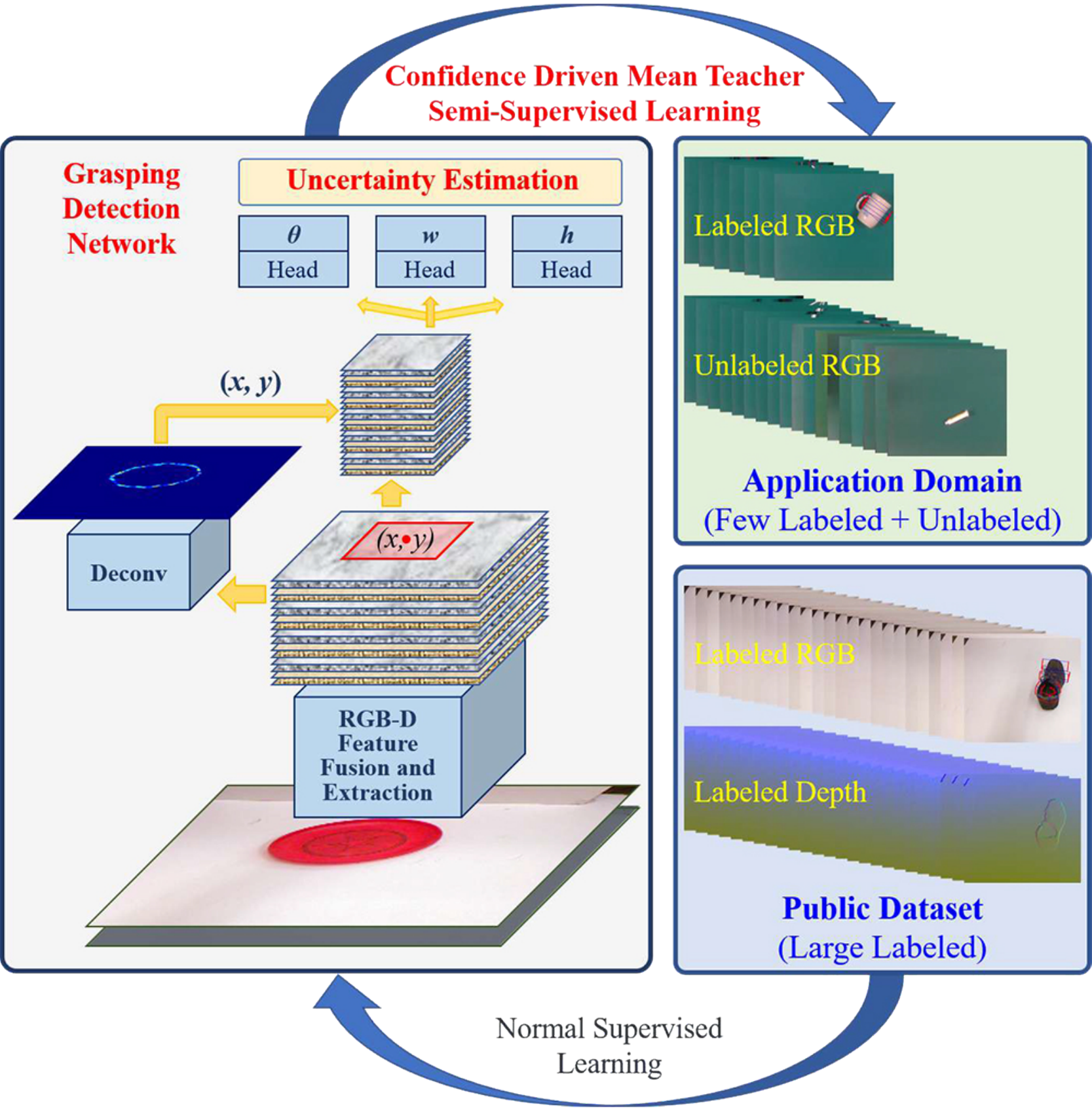}}
\caption{To facilitate the easy domain adaptation without tedious data labelling, a novel grasping detection network is proposed with a prediction uncertainty estimation mechanism. Furthermore, a confidence-driven mean teacher semi-supervised learning is used to adapt the domain by leveraging the cheap but massive unlabelled data with only a few labelled data.} \ \label{fig:Framework}
\end{figure}

In this work, we propose a novel detection architecture that separately treats the whole grasping problem as two subproblems, i.e., location (where) and pose (how), within one network. Different from the widely used region proposal approach~\cite{NIPS2015_Ren} in detection networks, the grasping location is inferred by a heatmap in our network through the encoder-decoder structure, which avoids the disadvantages of the anchor-based approaches such as the imbalance between positive and negative boxes, lots of hyperparameters~\cite{Law_2018_ECCV}, etc. The location heatmap is generalized from the limited grasping annotations and is able to indicate the grasping feasibility in the continuous space, which facilitates other high-level grasping planning if some additional requirements exist. In parallel, this architecture is also able to infer the optimal grasping pose (angle, width, and height) for every interested grasping location by using the local discriminative feature, thus it constitutes a full grasping detection in an integrated but flexible manner.

Specially, we implement an uncertainty estimation mechanism to evaluate the pose prediction in this architecture. Although Faster/Mask-RCNN also provides a confidence score for every detected box, such a score only represents whether an object is present within the anchor box from the classification view (softmax output), which actually provides no information about the box regression uncertainty. In this work, we propose an uncertainty estimation mechanism based on the Feature Pyramid Network (FPN). In each pyramid stage, the regression of the grasping pose is conducted. If a sample matches the trained distribution well and the prediction is certain, the variance of outputs from different pyramid stages is supposed to be small and vice versa. Moreover, by utilizing the uncertainty estimation mechanism, we propose a confidence-driven mean teacher semi-supervised learning for the grasping domain adaptation. The key idea is that, by leveraging on the cheap unlabelled data through semi-supervised learning, the model consistency is only emphasized for those pseudo-labelled data with high confidence (low uncertainty) instead of them all. This is able to prevent the model from learning the wrong consistency, so that it can improve and speed up the learning process. Overall, the whole picture is illustrated in Fig.~\ref{fig:Framework}.




The remainder of this paper is organized as follows. In Section II, we discuss the related work. Section III presents our approaches. The implementation details are introduced in Section IV. The experiments are presented in Section V, and finally, the conclusion is drawn in Section VI.

\section{Related Work}


\subsection{Deep Learning for Grasping Detection}

Visual learning approaches for grasping novel objects start in~\cite{Kamon1996,Ashutosh2008,Jiang2011,Bohg2014} by sampling and ranking candidate grasps. Deep learning is then introduced for multi-modal information such as RGB-D inputs. Sparse Auto-Encoder (SAE)~\cite{Lenz2015} or stacked SAE~\cite{Wang2016} with a two-stage approach are used, where a small network is to search a few candidates and a larger network is then to find the top-ranked rectangle. Single-stage methods are proposed~\cite{Redmon2015,Kumra2017} with the help of a deep network such as AlexNet~\cite{Alex2012}, ResNet~\cite{He_2016_CVPR}, etc., where the grasping detection is formulated as a regression problem to find one optimal location and pose. Multi-grasp detection networks are also proposed for grasping~\cite{Redmon2015,Chu2018,park2018realtime,Zhou2018}, inspired by the recent object detection methods like YOLO~\cite{Redmon_2016_CVPR} and Faster-RCNN~\cite{NIPS2015_Ren}, etc. The grasping orientation are also formulated as classification problem by discretization~\cite{Chu2018,Zhou2018}. Grasping detection network for object overlapping scenes are also proposed by finding the Region of Interest (ROI) first and then performing the grasping detection~\cite{zhang2018roibased}.

\subsection{Semi-Supervised Learning}

Semi-supervised learning exploits the unlabelled data to provide the regularization for reducing the model overfitting. It utilizes consistency regularization to make consistent predictions in response to the perturbation of unlabelled samples. Currently, most works are focused on classification. $\Pi$-model~\cite{Samuli16} evaluates the perturbed unlabelled samples twice using stochastic augmentation, dropout and Gaussian noise to minimize their difference and improve the consistency. Mean teacher~\cite{Antti17} offers a better teacher model for generating consistency targets through averaging model weights, and its student network is jointly trained with a standard supervised classification loss and an unsupervised consistency loss under different sample augmentations. However, minimizing consistency loss directly might be harmful as the pseudo targets generated by the teacher model are not the real ground truth and might be wrong. Emphasizing the consistency on such noisy data may mislead the training process and degrade the performance, especially when the labelled samples are scarce. In this work, we implement a detection network with prediction uncertainty estimation scheme, our confidence-driven mean teacher model try to leverage the uncertainty metric and only feed the selected confident samples for consistency loss minimization.

\subsection{Model Uncertainty Estimation}

The uncertainty of model prediction is traditionally modelled for the Bayesian neural network by integrating over the posterior distribution over parameters. Recently, it is theoretically proved that Dropout can be used to approximate a model's uncertainty, which can be considered as the Monte Carlo sampling from the posterior distribution of model~\cite{Gal_ICML_2016}. It utilizes the variance between multiple predictions with random dropout as a uncertainty metric. In literature, this uncertainty is commonly used for testing purpose only, which is not utilized during model training. In this work, we propose a prediction uncertainty metric based on the consistency between multiple predictions from different pyramid feature stages. Such an uncertainty metric is utilized to rank the detected grasps. More importantly, it is further used to improve the semi-supervised learning for our domain adaptation.

\section{Our Approach}

\begin{figure}[tb!]
\centering{\includegraphics[width=3.5in]{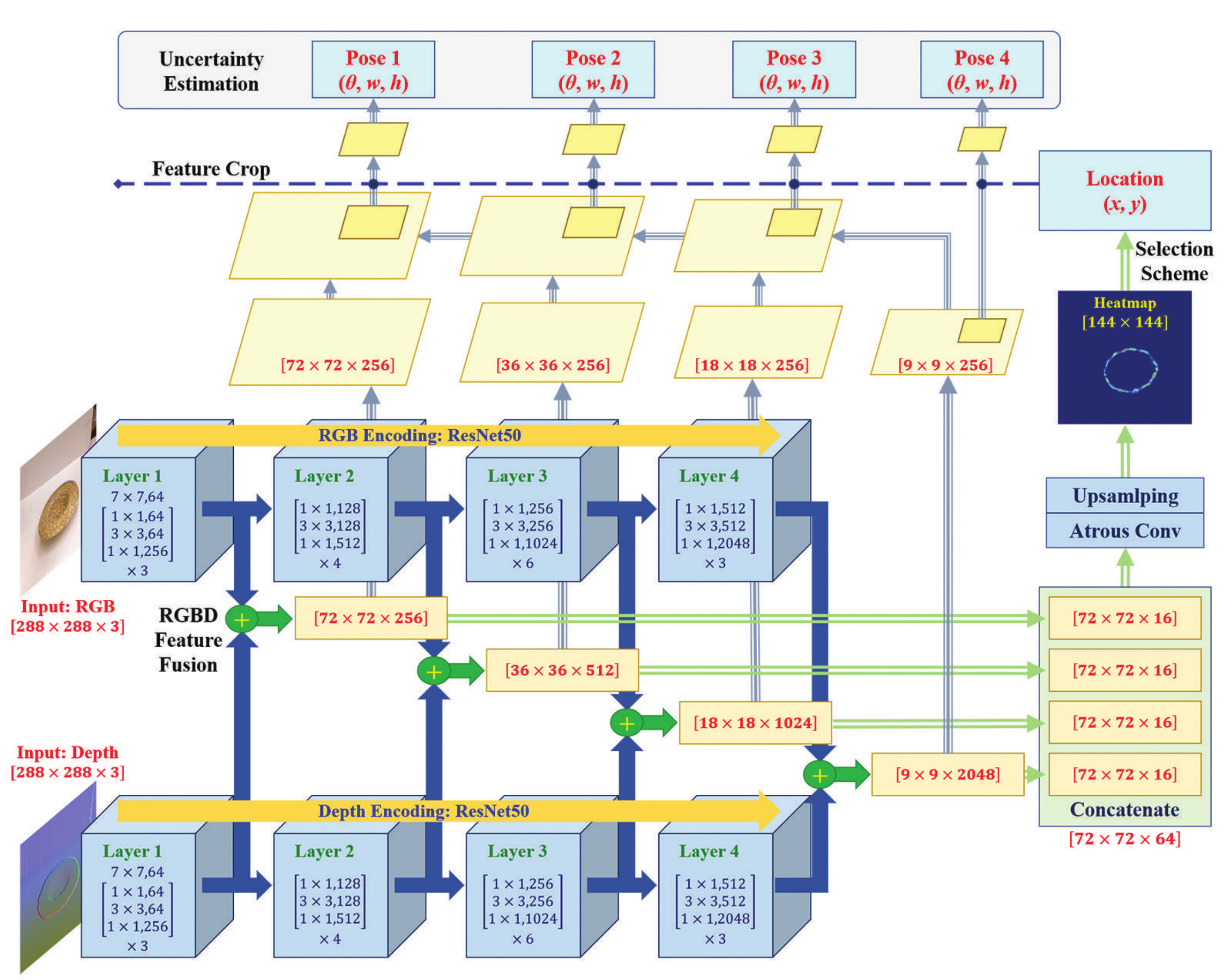}}
\caption{Grasping detection network architecture: RGB and depth are encoded in two separate branches by ResNet50 and fused on four different stages, which forms a pyramid feature network. One head decodes the pyramid feature and outputs a grasping location feasibility heatmap. The other four parallel heads crop the local feature around those interested locations on each pyramid feature map and then it predicts the grasping pose separately. The prediction uncertainty is estimated as the variance of four predicted poses.} \ \label{fig:architecture}
\end{figure}

The standard robotic grasping detection problem is formulated~\cite{Lenz2015} as given an RGB-D observation $\bm{o}$, design a network predictor $\bm{f}$ to predict the successful grasping rectangle $\bm{g}$, denoted as $\bm{g}=\bm{f(o)}$. The grasp rectangle $\bm{g}$ is represented with five parameters as
\begin{equation}\label{GraspingRectangle}
\begin{aligned}
\bm{g}=\big(x,\,y,\,\theta,\,w,\,h\big),
\end{aligned}
\end{equation}
where $\big(x,\,y\big)$ denotes the center of grasp rectangle, $\theta$ represents its orientation, $w$ denotes the width of gripper plate, and $h$ represents the gripper's opening distance. The grasping detection is a bit special as no real ground truth is fully available because the feasible grasping rectangles might be infinite while the annotations are always limited. 


In this work, we propose a grasping detection network to predict the multiple grasping rectangles with the estimated prediction uncertainty, where the network architecture is shown in Fig.~{\ref{fig:architecture}}. Generally speaking, this network treats the grasping detection as two subproblems: ``where" are the locations to grasp, denoted as LocNet that
\begin{equation}\label{LocNetDef}
\begin{aligned}
\big(x,\,y\big)\leftarrow\bm{m}=\bm{f_{l}}\big(\bm{o}\,|\,\bm{\phi_{s}},\bm{\phi_{l}}\big),
\end{aligned}
\end{equation}
and ``how" to grasp (referred as pose) for the shortlisted locations, denoted as PoseNet that
\begin{equation}\label{PoseNetDef}
\begin{aligned}
\big(\theta,\,w,\,h\big)=\bm{f_{p}}\big(\bm{o}\,|\,\bm{\phi_{s}},\bm{\phi_{p}},\,x,\,y\big),
\end{aligned}
\end{equation}
instead of directly treating it as a combined ``where\&how" problem through the overall judgement. Here, $\bm{m}$ is a grasping location feasibility heatmap, $\bm{\phi}$ denotes the network parameters, $\bm{f_{l}}$ and $\bm{f_{p}}$ share the common $\bm{\phi_{s}}$ for base feature extraction. Uniquely, the proposed network associates a pose prediction uncertainty metric for every $\big(\theta,\,w,\,h\,|\,\bm{o},\,x,\,y\big)$. As a result, the grasping detection network is supposed be more robust by choosing the low-uncertainty candidates to avoid the potential mispredictions. 

\subsection{RGB-D Fusion Based Feature Pyramid Network}

To better utilize the RGB and depth information, an RGB-D fusion based on FPN is utilized for feature extraction as depicted in Fig.~{\ref{fig:architecture}}. The base feature extraction network is commonly shared by both the ``where" and ``how" detectors $\bm{f_{l}}$ and $\bm{f_{p}}$, which is relied on ResNet-50 by naturally decomposing it into four block layers $\bm{l_{k}}$, $k=1$, 2, 3, and 4. For both the RGB and depth branches, the feature extraction is separately executed as
\begin{equation}\label{RGBDExtraction}
\begin{aligned}
\bm{x_{k}^{r}}=\bm{l_{k}^{r}}\big(\bm{x_{k-1}^{r}}\big),\
\bm{x_{k}^{d}}=\bm{l_{k}^{d}}\big(\bm{x_{k-1}^{d}}\big)
\end{aligned},
\end{equation}
where the subscript $r$ and $d$ denote the associations with RGB and depth branches, respectively, and $\big(\bm{x_{0}^{r}},\,\bm{x_{0}^{d}}\big)=\bm{o}$ is the input RGB-D pairs. The RGB-D feature fusion is achieved after each ResNet block layer as
\begin{equation}\label{RGBDFusion}
\begin{aligned}
\bm{x_{k}^{f}}=\bm{x_{k}^{r}}\oplus\bm{x_{k}^{d}},
\end{aligned}
\end{equation}
where $\oplus$ denotes element-wise summation.

\subsection{LocNet: Grasping Location Heatmap}


The proposed network uses an encoder-decoder Fully Convolutional Network (FCN) head to predict a feasibility heatmap for detecting the feasible/optimal grasping locations, where a higher heat indicates a higher grasping feasibility for this pixel $\big(x,\,y\big)$. For training this FCN head, its loss function is formulated as a binary classification problem for every pixel via the cross-entropy loss. Note that it is not possible to have the real ground truth $\bm{m^{t}}$ for all feasible grasping locations, in practical we treat every annotated grasp rectangle's center $\big(x,\,y\big)$ with its surrounding $r$-radius ball as the target $\bm{m^{t}}$. Although such a target map is discrete in every annotation, the advantage of using this encoder-decoder structure is that it is a generative detection approach. It can generalize from the sparse grasping location annotations and try to map out all the possible locations in a dense and continuous manner. Such information might also be useful in high-level task planning, e.g., differentiating the handler of a tool to grasp, etc., which is out the scope of this paper. For simplicity, the selection scheme for the obtained heatmap can be filtered by a threshold with the Non-Maximum Suppression (NMS) operation to finalize the shortlisted grasping locations.



\subsection{PoseNet: Grasping Pose with Uncertainty Estimation}
\subsubsection{Pyramid Pose Prediction}
With a list of shortlisted locations $\big(x,\,y\big)$ know where to grasp predicted by $\bm{f_{l}}$, another head of the proposed network will predict their corresponding pose $\big(\theta,\,w,\,h\big)$ to further determine how to grasp based on the local discriminative features $\bm{x_{k}^{c}}$. This is a fixed-size sub-feature map around the shortlisted locations $\big(x,\,y\big)$ for every pyramid feature layer $\bm{x_{k}^{p}}$, represented as
\begin{equation}\label{Crop}
\begin{aligned}
\bm{x_{k}^{c}}=\bm{Cr}\Big(\bm{x_{k}^{p}},\,\big(x,\,y\big)\Big),
\end{aligned}
\end{equation}
where $\bm{Cr}$ denotes the crop operation and $k=1$, 2, 3, and 4. Consequently, four channels of individual fully connected network regression heads $\bm{f_{ph}^{k}}$ are utilized to predict the grasping pose on each pyramid feature layer as,
\begin{equation}\label{PoseInf}
\begin{aligned}
\big(\theta_{k},\,w_{k},\,h_{k}\big|\,x,\,y\big)&=\bm{f_{ph}^{k}}\big(\bm{x_{k}^{c}}\big)=\bm{f_{ph}^{k}}\Big(\bm{Cr}\Big(\bm{x_{k}^{p}},\,\big(x,\,y\big)\Big)\Big),
\end{aligned}
\end{equation}
and finally for a given location $\big(x,\,y\big)$, the mean value is treated as the pose prediction,
\begin{equation}\label{PosePred}
\begin{aligned}
\big(\theta,\,w,\,h\big)=\emph{mean}\big(\theta_{k},\,w_{k},\,h_{k}\big)\big|_{k=1,2,3,4}.
\end{aligned}
\end{equation}

\subsubsection{Uncertainty Estimation}
The proposed pyramid pose detection head enables a special mechanism to estimate the pose $\big(\theta,\,w,\,h\big)$ prediction uncertainty for every $\big(x,\,y\big)$. This is based on the common sense that for a well-trained pose prediction, the prediction results from different pyramid stages should agree with each other, and large variance leads to high uncertainty prediction. As a result, the variance among the predictions from all four pyramid heads is utilized as the uncertainty metric for the pose prediction at $\big(x,\,y\big)$, denoted as $M_{uc}\big(x,\,y\big)$ that,
\begin{equation}\label{Muc}
\begin{aligned}
M_{uc}\big(x,\,y\big)=\sum_{\tau=\theta,\,w,\,h}\emph{var}\big(\tau_{k}\big)\big|_{k=1,2,3,4}.
\end{aligned}
\end{equation}
For the shortlisted $\big(x,\,y\big)$ locations, $M_{uc}\big(x,\,y\big)$ helps to identify the most confident predictions of $\big(\theta,\,w,\,h\,|\,x,\,y\big)$, which can be ranked as the optimal grasping candidates. 

\subsection{Confidence-Driven Mean Teacher}

To achieve domain adaptation via only a small labelled data, the mean teacher semi-supervised learning is adopted in this work. The original mean teacher method is proposed for the classification tasks, we extend its usage on the detection regression network with our new contribution in prediction uncertainty filtering, which we refer it as confidence-driven mean teacher learning. The mean teacher method makes use of the consistency loss to prevent overfitting to limited labelled data. The student learns from the soft pseudo targets provided by the teacher for those unlabelled samples. However, such a pseudo target from the teacher can be wrong itself, the reinforcing on wrong targets will be harmful to the student as it deviates the model convergency. The motivation of confidence-driven mean teacher learning is to let the student only learn those pseudo targets with low uncertainties, which are supposed to be more correct targets. As the training process goes on, the model will become more certain in the new domain, so that more and more pseudo targets with low uncertainties will be added into the training pool gradually for consistency loss optimization. The student model will be benefited from learning those more correct prediction's consistency instead of emphasizing random consistency, thus it speeds up the training progress and improves the overall accuracy.

Let a training set $D_{t}$ consists of $N_l$ labelled samples and $N_u$ unlabelled samples, denoted as $D_{l}=\big\{(\bm{e_{i}}, \bm{t_{i}})\big\}_{i=1}^{N_l}$ and $D_{u}=\big\{\bm{e_{i}}\big\}_{i=1}^{N_u}$, respectively, and $D_{t}=D_{l}\cup D_{u}$. Define a teacher model $\bm{f^{t}(\phi'})$ and a student model $\bm{f^{s}(\phi})$, these two models share the same network structure but the parameters of teacher model is updated from the student model in every step $k$
\begin{equation}\label{ParaUpdate}
\begin{aligned}
\bm{\phi'_{k}}=\alpha \bm{\phi'_{k-1}}+(1-\alpha)\bm{\phi_{k}},
\end{aligned}
\end{equation}
where $\alpha$ is a smoothing coefficient hyperparameter. For every training batch $B$, it optimizes the student model by the minimization the combination of supervision loss and consistency loss
\begin{equation}\label{SemiLoss}
\begin{aligned}
\min_{\bm{\phi}}\Big\{&\,\sum_{\bm{e_{i}}\in D_{l}}\mathcal{L}\big(\bm{f^{s}}(\bm{e_{i},\,\phi,\,\mu}),\bm{t_{i}}\big)\\
&+\sum_{\bm{e_{i}}\in \overline{D}_{u}}\mathcal{L}\big(\bm{f^{t}}(\bm{e_{i},\,\phi',\,\mu'})-\bm{f^{s}}(\bm{e_{i},\,\phi,\,\mu})\big)\Big\},
\end{aligned}
\end{equation}
where $\bm{\mu}$ and $\bm{\mu'}$ denote the perturbation parameters for the teacher and student models, e.g., augmentation, etc. The uniqueness of our confidence-driven mean teacher exists on
\begin{equation}\label{DSub}
\begin{aligned}
\overline{D}_{u}=\big\{\bm{e_{i}}\,\big|\,\bm{e_{i}}\in{D}_{u}\,\&\,M_{uc}\big(\bm{f^{t}}(\bm{e_{i}})\big)<\overline{T}\big\},
\end{aligned}
\end{equation}
which is a subset of ${D}_{u}$ that only the samples with low prediction uncertainty ($M_{uc}\big(\bm{f^{t}}(\bm{e_{i}})\big)<\overline{T}$) will be filtered into $\overline{D}_{u}$, and $\overline{T}$ is a threshold hyperparameter. 

\section{Implementation Details}

\subsection{Grasping Dataset and Preprocessing}

We use the public Cornell grasping dataset (RGB-D) for training the initial grasping detection network. For evaluation of domain adaptation using the proposed confidence-driven mean teacher learning, we also collect a small labelled+unlabelled dataset including 360 images for 18 objects, and only 90 images are labelled with ground truth grasps (five per object). The small new domain dataset is acquired by Intel RealSense D435 RGB-D camera. However, the depth is quite noisy due to the small depth change. Therefore, the small dataset is provided with RGB images only, and the objects, view angle, and background are different from the Cornell grasping dataset for the evaluation of domain adaptation. The original images in both datasets are with 640$\times$480 resolution, they are resized into 456$\times$342 resolution and then cropped into 288$\times$288 as the input to grasping detection network. The depth images are processed into three channels by combining the original channel with Sobel filtered channels in $x$ and $y$. 

\subsection{Training of Grasping Detection Network}

The training process of the grasping detection network evolves two steps that the PoseNet and LocNet are trained separately and subsequently. For training the PoseNet, the network is fed with the RGB-D pairs and locations of the grasps in a batch of 16, and then predicts the poses $\big(\theta,\,w,\,h\big)$ for all pyramid stages. The PoseNet is supervised by minimizing the total losses for all pyramid stages' predictions. The initial learning rate is set as 0.0001 and then gradually decreases, the SmoothL1Loss is used for PoseNet training. Subsequently, the learned coefficients of ResNet-50 layers in PoseNet are passed to LocNet and then keep fixed. The remaining layers of LocNet are further trained by feeding the RGB-D pairs and the targets as described in Section III B, where the CrossEntropyLoss is used and the initial learning rate is 0.0003. The reason we train PoseNet first with ResNet coefficients and then fixed for LocNet is that grasping pose is more important than the location heatmap as we have the certainty based ranking scheme to fine select the optimal location. After this two-step training, the learned coefficients of LocNet and PoseNet are reloaded to the full grasping detection network which can perform the grasping detection tasks with prediction certainty estimation.

\begin{figure}[tb!]
\centering{\includegraphics[width=3.5in]{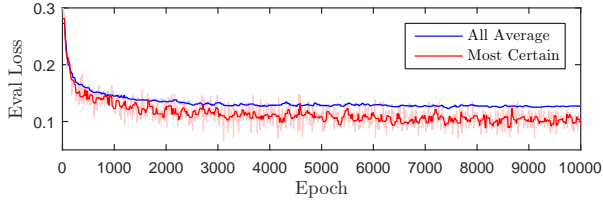}}
\caption{Comparison between the All Average prediction error against the Most Certain prediction error on the evaluation set.} \ \label{fig:AvVsMc}
\end{figure}

\subsection{Semi-Supervised Domain Adaptation}

We apply the proposed confidence-driven mean teacher learning for domain adaptation that enables the grasping detection network learned from the public dataset to perform better for the new application domain with only a small labelled data. The evaluation set contains 54 labelled samples (3 per object) for evaluation. Initially, the LocNet is fine-tuned directly by fully supervised learning using 18 labelled samples. The semi-supervised learning is applied to the head of PoseNet while the ResNet-50 layers keep fixed. For student model PoseNet, the training batch consisting of 6 labelled samples and 2 pseudo labelled samples is fed iteratively to minimize the combined loss, where the perturbation effect is from the sample augmentation. In each step, the teacher model is updated from the new student model, where $\alpha$ is gradually increased from 0.5 to 0.99.

\section{Experiments and Discussions}

\subsection{Grasping Detection Network with Certainty Estimation}

\begin{figure}[tb!]
\centering{\includegraphics[width=3.4in]{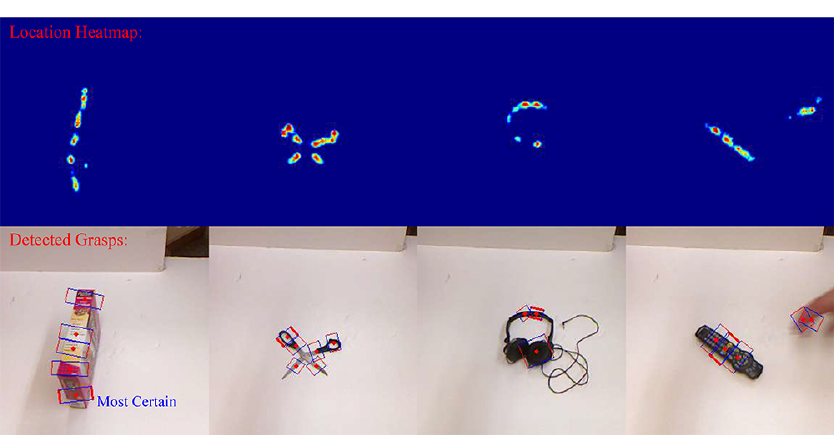}}
\caption{Predicted location heatmap (above) and grasps (below).} \ \label{fig:MultiGrasp}
\end{figure}

The evaluation of the proposed grasping detection network is conducted on the Cornell grasping dataset. We first evaluate our proposed uncertain estimation scheme that the lowest-uncertain prediction outperforms the other cases. Fig.~\ref{fig:AvVsMc} shows that the average of the most certain prediction errors against the average of all prediction errors during the whole training progress of PoseNet, where the recorded loss is from the evaluation set that never used for training. It can be seen that the most certain cases are generally better than the average cases around 30\% in the loss. Fig.~\ref{fig:MultiGrasp} shows the prediction results including location heatmap and detected grasps using the proposed network.

\begin{figure}[tb!]
\centering{\includegraphics[width=3.4in]{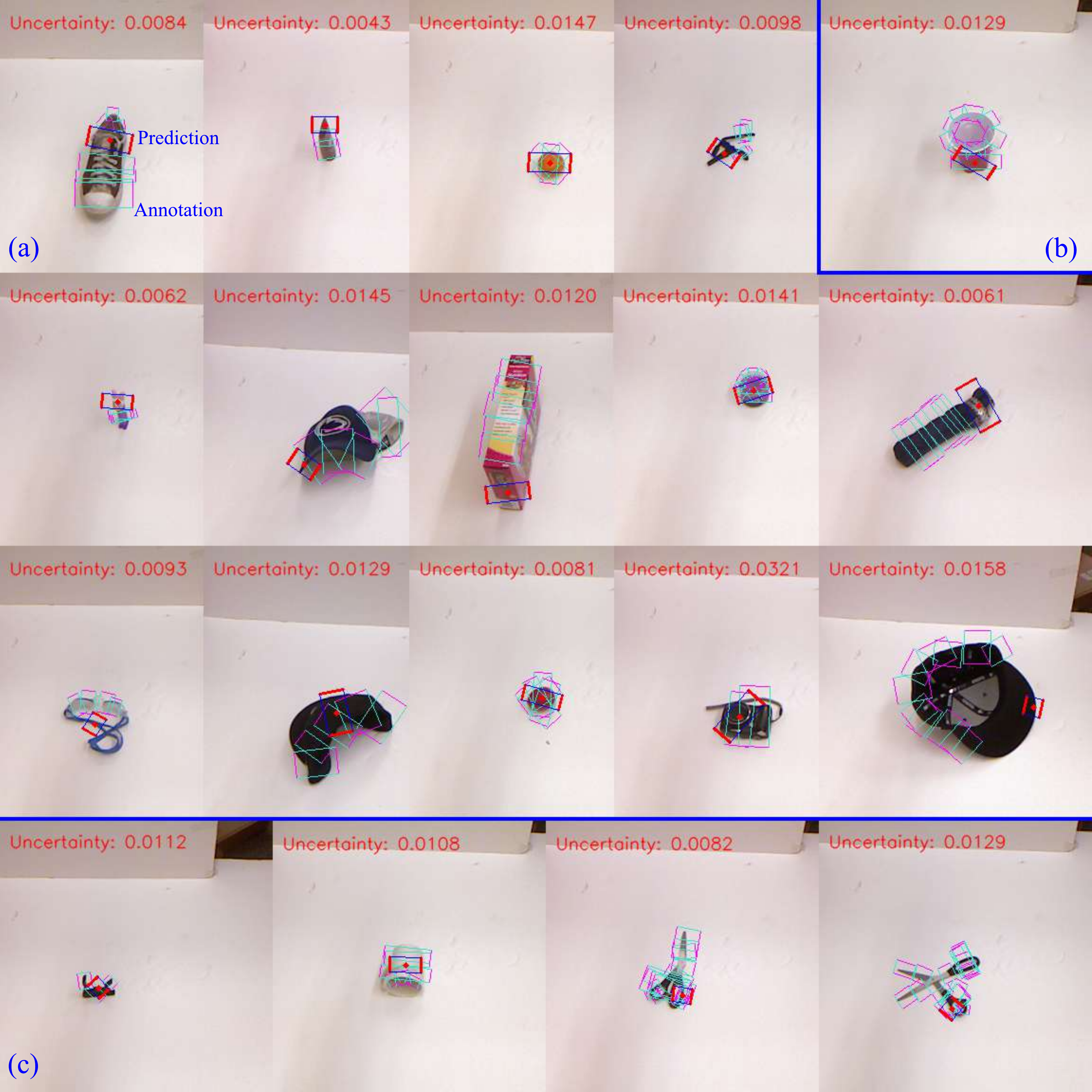}}
\caption{Evaluation on Cornell grasping dataset: (a) 14 False Negative samples, (b) 1 True Negative sample, and (c) 4 False Positive samples.} \ \label{fig:CornellSet}
\end{figure}

The commonly used grasping evaluation metric is that a candidate grasp is viewed as correct if 1) the difference of angle between predicted grasp $\bm{g_p}$ and ground truth $\bm{g_t}$ is within $30^\circ$, and 2) the Intersection over Union (IoU) of the predicted grasp $\bm{g_p}$ and the ground truth $\bm{g_t}$ is greater than 0.25. Base on this criterion, the proposed grasping detection network achieves 92.2\% (177/192) accuracy in the test set (192 samples, 20\% image-wise split), where the lowest-uncertain grasp is used for evaluation. However, the above criterion may not able to reflect the real success rate of the grasping as it only accounts for the IoU and angle difference, while the collision between the gripper and object is not considered. By manually analyzing the predictions of all 15 negative samples, we found that 14 cases of them are actually false-negative and only 1 sample is true-negative, shown as in Fig.~\ref{fig:CornellSet} (a) and (b), respectively. 4 false-positive predictions are found among 177 positive samples, as shown in Fig.~\ref{fig:CornellSet} (c). Therefore, the real success rate is 187/192=97.4\%.



\subsection{Confidence-Driven Mean Teacher Domain Adaptation}

\begin{figure}[tb!]
\centering{\includegraphics[width=3.45in]{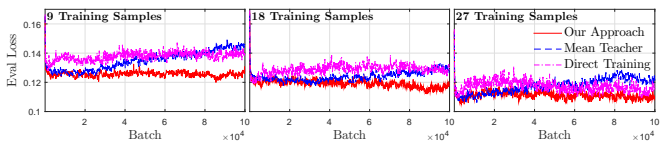}}
\caption{Comparison of average loss on evaluation set by using (a) 9, (b) 18, and (c) 27 labelled samples for training.} \ \label{fig:SemiComp}
\end{figure}

To evaluate the domain adaptation using only limited data, we test the algorithms using 9, 18, and 27 labelled samples for PoseNet, respectively. Fig.~\ref{fig:SemiComp} plots their comparisons of average loss on a new domain evaluation set. Three training methods are implemented using the same setting, i.e. direct training (fine-tuning) with those labelled data, the original mean teacher to minimize consistency loss for all unlabelled data, and our proposed confidence-driven mean teacher that only minimizes consistency loss for these confident samples with low uncertainties. It can be seen from Fig.~\ref{fig:SemiComp} that the confidence-driven mean teacher can significantly prevent the overfitting in all three cases, while the direct training suffers from the overfitting shortly due to the small number of training sets. Compared with the original mean teacher, the learning speed is faster in the confidence-driven mean teacher, which is because the original mean teacher learns much noisier information with incorrect labels. Moreover, although the original mean teacher performs better than the direct training at the beginning (same level with confidence-driven mean teacher), it can be seen that it disastrously diverges as the training goes on for all three cases, where the reasons are supposed to be same as above due to the emphasizing on the consistency of the wrong pseudo targets.

\begin{figure}[tb!]
\centering{\includegraphics[width=3.4in]{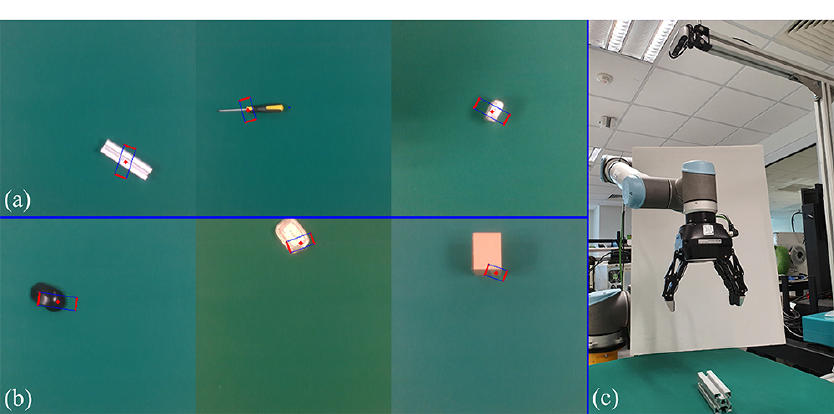}}
\caption{Prediction evaluation on new domain: (a) correct predictions, (b) incorrect predictions, and (c) the real grasping testbed.} \ \label{fig:DomainImg}
\end{figure}

It is also noted that for real new domain adaptation applications, it is much desired to not have a large labelled evaluation set for having such plots like Fig.~\ref{fig:SemiComp}, all labelled data should be fully used in training to maximize the accuracy. In such a scenario, the confidence-driven mean teacher provides more robust performance in overcoming the overfitting and model diverging. Statistically, by using only 9 labelled data for domain adaptation, the success rate is 145/162=89.5\% for adapting from RGB-D to only RGB with different backgrounds, etc. Finally, the grasping detection results using 9 labelled data for domain adaptation are plotted in Fig.~\ref{fig:DomainImg} with both correct cases (a) and incorrect cases (b), and the network is implemented on the UR robot with Robotiq gripper where the testbed is shown as in Fig.~\ref{fig:DomainImg}(c).

\section{Conclusions}

This paper addresses the data-efficient domain adaptation for the robotic grasping detection, where only a few labelled data is desired for real applications to minimize the tedious labelling work. We present a grasping detection network with prediction uncertainty estimation and a confidence-driven mean teacher semi-supervised learning algorithm, these two parts are closely interacted to form an integrated solution for data-efficient domain adaptation. Our results show that the proposed detection network is able to perform the grasping detection in high accuracy, and the confidence-driven mean teacher outperforms the original mean teacher and direct training for the detection regression tasks in avoiding the overfitting and model diverging.

\addtolength{\textheight}{-2cm}   




\appendices

\bibliographystyle{IEEEtran}

\bibliography{Deep_Grasping_Detection}

\end{document}